\documentclass[journal,onecolumn]{IEEEtran}
\usepackage{algorithm}
\usepackage{algorithmic}
\usepackage{multirow}
\usepackage{lineno,hyperref}
\usepackage{xcolor}
\usepackage{graphicx}
\usepackage{booktabs,comment,balance}
\begin{document}
%
\title{
Quantification of Occlusion Handling Capability of a 3D Human Pose Estimation Framework
}

%
%
%

 \author{Mehwish~Ghafoor,
         Arif~Mahmood
 \thanks{M Ghafoor and A Mahmood are with the Department of Computer Science, Information Technology University, Lahore, Pakistan, emails: \{mehwish.ghafoor, arif.mahmood\}@itu.edu.pk}
\thanks{Manuscript received July 19, 2021,This is a draft of the paper accepted in IEEE Transaction on Multimedia, 2022}}

\markboth{IEEE TRANSACTIONS ON MULTIMEDIA, July~2021}%
{Ghafoor \MakeLowercase{\textit{et al.}}: Quantification of Occlusion Handling Capability of 3D Human Pose Estimation Framework }

\maketitle

\begin{abstract}
3D human pose estimation using monocular images is an important yet challenging task. 
Existing 3D pose detection methods exhibit excellent performance under normal conditions however their performance may degrade due to occlusion. Recently some occlusion aware methods have also been proposed, however, the occlusion handling capability of these networks has not yet been thoroughly investigated.
In the current work, we propose an occlusion-guided 3D human pose estimation framework and quantify its occlusion handling capability by using different protocols. The proposed method estimates more accurate 3D human poses using 2D skeletons with missing joints as input. Missing joints are handled by introducing occlusion guidance that provides extra information about the absence or presence of a joint. Temporal information has also been exploited to better estimate the missing joints. A large number of experiments are performed for the quantification of occlusion handling capability of the proposed method on three publicly available datasets in various settings including random missing joints, fixed body parts missing, and complete frames missing, using mean per joint position error criterion. 
In addition to that, the quality of the predicted 3D poses is also evaluated using action classification performance as a criterion. 3D poses estimated by the proposed method achieved significantly improved action recognition performance in the presence of missing joints. Our experiments demonstrate the effectiveness of the proposed framework for handling the missing joints as well as quantification of the occlusion handling capability of the deep neural networks.
\end{abstract}


%
\IEEEpeerreviewmaketitle

\section{Introduction}
\label{intro}
 \IEEEPARstart{3D} {human} pose estimation is an important problem having wide range of applications including video surveillance \cite{das2020vpn}, action recognition~\cite{angelini20192d}, avatar creation, motion pasting~\cite{weng2019photo}, anomaly detection \cite{markovitz2020graph},  tracking~\cite{bao2020pose,li2019multi}, and in healthcare~\cite{srivastav2020self}. Automatic 3D pose estimation is a challenging problem because of non-rigid human structure consisting of many joints and relatively rigid limbs. Therefore, 3D pose estimation is often referred to as estimating positions of various joints in 3D space. In case of partial or complete occlusion caused by other objects or humans in the scene, some or all joints may get missing or get degraded confidence which exacerbates the difficulty of 3D pose estimation problem and degrades the performance of the downstream applications using detected 3D pose as input.
 

 Due to numerous 3D human pose applications, in the recent years many researchers have focused on this area. Many 
 3D pose estimation methods leverage the strength of 2D detectors by using detected 2D joints as an input \cite{rayat2018exploiting,chen2019unsupervised,pavllo20193d,zhao2019semantic,liu2020attention}. Most of these methods face difficulty in recovering accurate 3D pose if the subject is partially or completely occluded. It is because most of these methods do not explicitly handle occlusion \cite{pavllo20193d,zhao2019semantic,liu2020attention}. Recently some occlusion aware pose detection methods have also been proposed, however, the occlusion handling capability of these methods have not been thoroughly quantified \cite{cheng2019occlusion,cheng20203d,gu2021exploring,qammaz2021occlusion}. Some methods show improved performance on the publicly available datasets when occlusion aware training was used~\cite{cheng2019occlusion,cheng20203d}, while others report very few experiments~\cite{qammaz2021occlusion,moreno20173d,a02}. In contrast to the existing approaches, we propose a method to explicitly handle missing joints and we quantify the occlusion handling capability of the proposed method. We compare it with the existing state-of-the-art methods in terms of Mean Per Joint Position Error (MPJPE) as well as by using the accuracy of action recognition as a criterion.
\begin{figure}[t]
\centering
\includegraphics[width=0.45\textwidth]{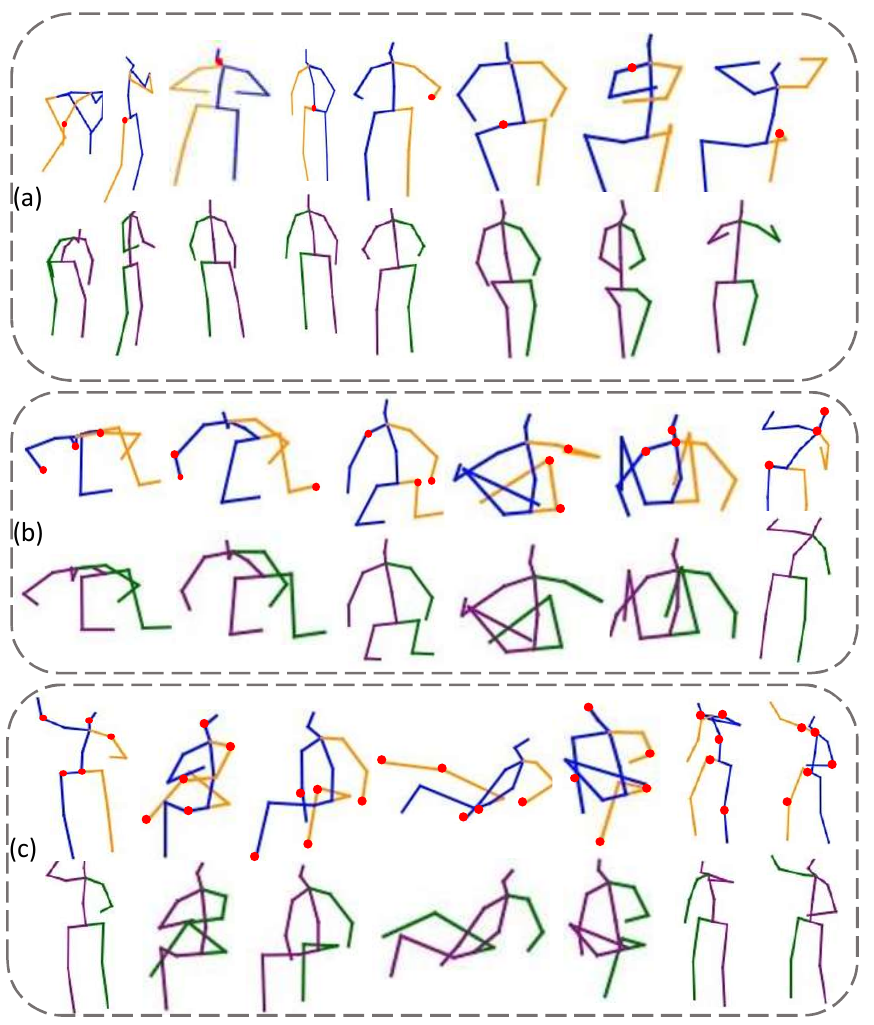}
\caption{3D human pose estimation by the proposed method in Human 3.6M dataset using (a) only 1 out of 17, (b) 3 out of 17, (c) 5 out of 17 random 2D input joints. In each case the top row shows ground-truth and the bottom row shows the estimated 3D poses.}
\label{fig:input_seq}
\end{figure}

The proposed occlusion guided 3D human pose estimation framework is based on a temporal dilated CNN which can estimate more accurate 3D joint positions even in the presence of severe occlusion. Explicit occlusion guidance is obtained by using an indicator variable which gives the additional information to the network regarding the presence or absence of a joint. The occlusion handling capability of the proposed method is quantified by randomly missing up to $n_p-1$ joints, where $n_p$ are the total joints in the pose (keeping only one random joint in each frame). Experiments are also performed by missing all joints in 1, 3 and 5 consecutive frames in a sequence. In addition to that, experiments are also performed by missing fixed body parts such as missing lower body, left arm, etc. Performance comparisons with the existing state of the art methods reveal significant improvement in the missing joint estimation in terms of MPJPE. 
Moreover, we also quantify the occlusion handling capability of the proposed method as well as existing state of the art methods by using 3D pose based action recognition as the target application. To this end, we use a baseline method as well as a recent Graph Convolutional Neural network (GCN) based method~\cite{zhang2020semantics}. 
We observe significant action recognition performance improvement when 3D poses were estimated using the proposed occlusion guided framework. Numerous experiments are performed for occlusion handling quantification on three datasets including Human $3.6$M \cite{human3}, NTU RGB+D~\cite{shahroudy2016ntu} and  SYSU~\cite{hu2015jointly} and compared with five existing state of the art methods. In both types of quantification experiments, proposed framework has exhibited significant performance improvement in terms of MPJPE as well as action recognition accuracy. 
\begin{figure*}[!t]
\centering
\includegraphics[width=\textwidth]{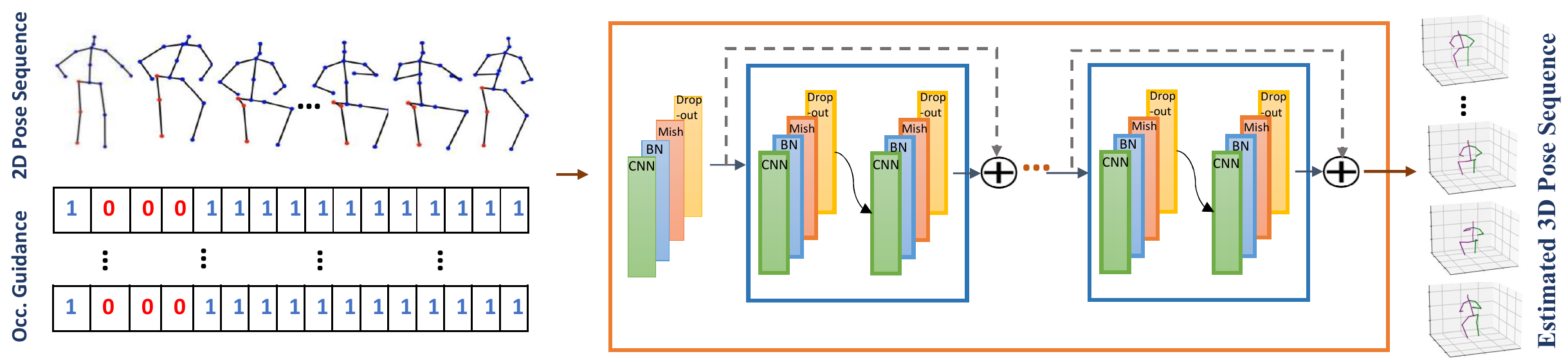}
\caption{Proposed occlusion guided 3D pose estimation framework is based on a temporal dilated CNN with residual connections.}
\label{fig:teaser}
\end{figure*}
The main contributions of the current work are as follows:
\begin{enumerate}
    \item An occlusion guided framework is proposed based on temporal dilated CNN for the estimation of 3D human body pose in the presence of severe occlusions. 
    \item Occlusion handling capability of deep neural networks is quantified by randomly missing 2D input joints, missing fixed body parts and missing all joints in few frames of a sequence. The quality of estimated poses is quantified using the mean per joint position error as well as action recognition accuracy. 
    \item Comprehensive evaluations on three publicly available datasets including Human $3.6$M, NTU RGB+D and SYSU have shown significant improvement in both 3D joint estimation and action recognition accuracy.
\end{enumerate}

The rest of this paper is organized as follows: Section \ref{sec:related} presents the related work, the proposed approach and experimental results are presented in Sections \ref{sec:classifiers} and \ref{sec:exp & res} respectively, followed by conclusions and future directions in Section \ref{sec:con}.
\vspace{-3mm}
\section{Related work} 
\label{sec:related}
3D human pose estimation has received much attention by the research community and several new directions have emerged \cite{rayat2018exploiting,chen2019unsupervised,pavllo20193d,liu2020attention,lee2018propagating,wandt2019repnet,chen2020anatomy,nie20203d}. A significant number of detectors exploit the much-developed 2D methods for 3D pose estimation. Martinez \textit{et al.}~\cite{a01} introduced a fully connected neural network with two residual modules to estimate 3D pose using 2D joint positions as input. Fang \textit{et al.}~\cite{fang2018learning},~\cite{xu2021monocular} improved 3D pose estimation using pose grammar network. Cheng \textit{et al.}~\cite{Cheng2021GraphAT} proposed directed joint and bone based Graph Convolutional Network (GCN) for 3D pose estimation in multi-person scenario. Most of these methods get static pose as input and therefore do not use temporal information.
{Some later methods exploit the temporal information for 3D pose estimation. Hossain \textit{et al.}~\cite{rayat2018exploiting} used LSTM seq-to-seq architecture to ensure temporal smoothness, however its limitation is fixed-size input and output. Later, Pavllo \textit{et al.}~\cite{pavllo20193d} introduced temporal dilated convolution to estimate 3D pose from sequence of 2D poses and maintained efficiency with dilated convolution. Attention mechanism is also incorporated in temporal convolution to get improved results \cite{liu2020attention}. Most of these methods do not provide any occlusion handling mechanism.}
\begin{figure}[b]
\centering
\includegraphics[width=0.5\textwidth]{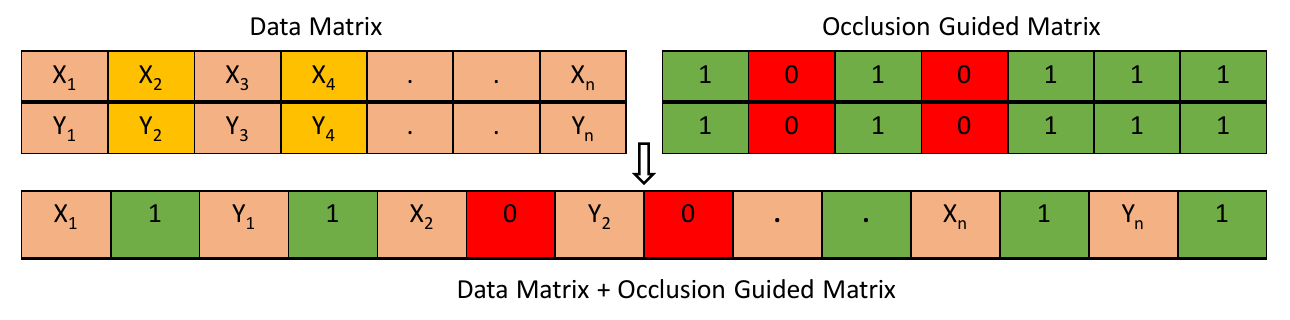}
\caption{Proposed occlusion guidance mechanism  provides precise information about missing 2D joints. }
\label{fig:2}
\end{figure}

Some researchers have recently addressed the occlusion handling challenge in 3D pose estimation.
Moreno Noguer~\cite{moreno20173d} proposed a static 3D pose estimation method by representing 2D and 3D poses as Euclidean distance matrices which are robust to occlusion. Park \textit{et al.}~\cite{a02} proposed a static relational hierarchical dropout method (Rel-Hier-Drop) to handle occlusion and noisy 2D poses. Cheng \textit{et al.}~\cite{cheng2019occlusion} used cylinder man model for data augmentation to train the network in occlusion aware fashion using a sequence of RGB video frames.
Cheng \textit{et al.}~\cite{cheng20203d} used training data augmentation with partial and complete occlusion. For reliable 3D pose estimation, spatial and temporal kinematic chain space discriminator is also used. Both methods \cite{cheng2019occlusion,cheng20203d} have obtained improved results on publicly available datasets with little occlusion. However, to quantify the capability of their networks for estimation of missing joints, no occlusion handling results are reported. Zhang \textit{et al.}~\cite{zhang2020object} estimated 3D pose and reconstructed occluded human body shape using 3D mesh model. Qammaz \textit{et al.}~\cite{qammaz2021occlusion} proposed normalized signed rotation matrices that are translation and scale invariant and occlusion tolerant to classify the orientation of 2D pose. 3D pose is then estimated using ensemble of neural networks. Gu \textit{et al.}~\cite{gu2021exploring} used soft gated CNN for 3D pose estimation which acts as attention to handle occlusion from noisy 2D joints.

In contrast to the existing state-of-the-art occlusion handling approaches, we propose occlusion guided method based on temporal dilated CNN for the estimation of 3D poses. In large number of experiments, we quantify the capability of our network for missing joints estimation by randomly occluding up to 94\% in Human 3.6M, up to 90\% joints in SYSU, and up to 92\% joints in NTU RGB+D dataset. We observe significant performance improvement in MPJPE in the presence of large number of missing joints, as well as improvement in action recognition performance. In the current work, we use action recognition as a quality measure of estimated 3D poses. As the number of missing 2D joints increases, the quality of estimated 3D poses degrades resulting in the loss of action recognition performance. Our proposed occlusion guided 3D pose estimation framework has shown a graceful degradation with increasing number of missing joints.   




\begin{figure*}[t]
\centering
\includegraphics[width=0.9\textwidth,keepaspectratio]{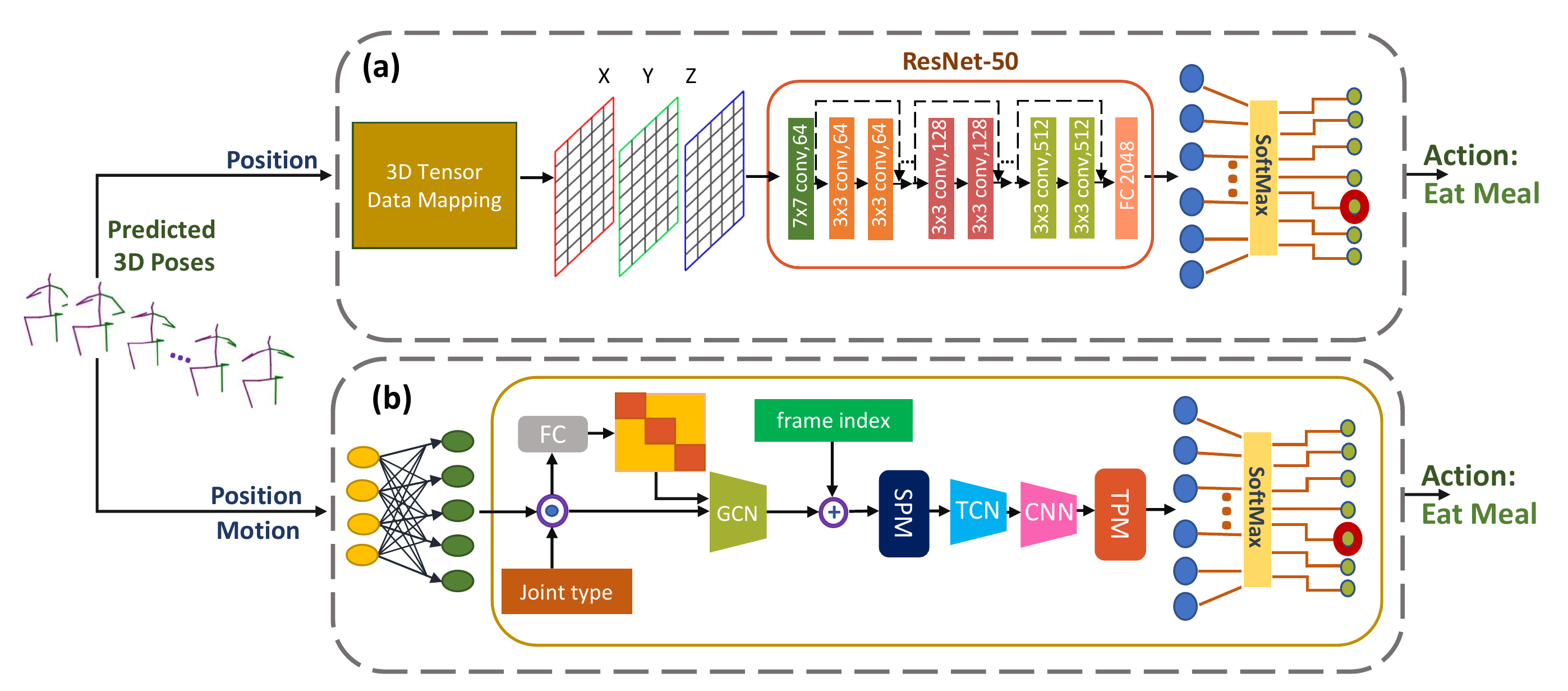}
\caption{Action recognition framework: (a) Predicted 3D Poses are reshaped as 3D tensor and input to the baseline CNN model. (b) Predicted 3D Poses are input to a GCN with spatial and temporal positions of joints.}
\label{fig:action}
\end{figure*}

\section{Proposed Occlusion Guided 3D Pose Estimation Framework}
\label{sec:classifiers}
The proposed framework as shown in Fig.~\ref{fig:teaser} consists of a temporal dilated CNN with an occlusion guidance mechanism to handle the missing joints.  

\vspace{-4mm}
\subsection{Temporal Dilated CNN}\label{gen}
The proposed network is a fully convolutional architecture with residual connections which takes $f$ consecutive 2D poses as input and estimates a 3D pose corresponding to the central location of the input sequence. 
The proposed architecture consists of multiple CNN blocks. Regular convolution is replaced  with dilated convolution to allow larger receptive fields with lower parameter cost and higher efficiency without any resolution compromise. The input layer of the framework takes $n_p\times d \times f$ temporal sequence and outputs $1024$ channels, where $n_p$ is the number of joints in one skeleton, $d$ is the dimensions of each joint including occlusion guidance matrix, and $f$ are the number of consecutive skeletons in the temporal sequence. The network estimates $n_p \times d$ 3D pose, where $d=3$ because the estimated pose is 3D.
Each block consists of two CNN layers where $1^{st}$ layer consists of 1D CNN with dilation of exponential factor $D=\{3, 9, 27, 81, 243\}$, Batch Normalization (BN), Mish activation function \cite{mish2019self}, and dropout followed by the second layer which consists of 1D CNN  with standard convolution (D=1), BN, Mish and dropout layers. The proposed framework consists of four such blocks with residual connections (see Table \ref{ta:np}). In the output layer, $1024$ channels shrink to $n_p\times d$ to get the required dimension using temporal information from past and future frames. Following objective function is minimized while training:
\begin{equation}
\mathcal{L} = \frac{1}{n_b}\sum_{i \in n_b} ||p_3^i-\widehat{p}_3^i||_2
\end{equation}
where $n_b$ indicates batch size, $p_3^i$ and $\widehat{p}_3^i$ are the ground-truth and the predicted 3D poses, respectively.


\vspace{-3mm}
\subsection{Occlusion Guidance Mechanism}\label{ind}
Most of the existing 2D pose detectors also generate a confidence score for each joint in addition to the geometric joint position. In case of occlusion or failure of joint estimation the value of confidence score degrades, depicting the quality loss of estimated joint. In order to avoid incorrect data input for the estimation of 3D pose, joints with low confidence score may be considered as missing . To effectively handle these missing joints, an occlusion guidance matrix is introduced which has a value of 1 for available joints (or high confidence joints) and a value 0 for the missing joints (low confidence joints). The input to the framework consists of geometric 2D joint positions and two indicator variables corresponding to both coordinates as shown in Fig. \ref{fig:2}. The occlusion guidance methodology provides missing data information to the network. Both the indicator variable and corresponding coordinates are coupled together, effectively doubling the input dimensionality. The proposed T3D CNN estimates missing values as well as 3D pose simultaneously. 

\begin{table}[b]
\small
\centering
\caption{Number of parameters in Millions(M) in proposed T3D CNN. }
\begin{tabular}{|c|c|c|c|c|c|}
\hline
Layer $\#$ &{1-2}& {3-4} & {5-6} & {7-8} & {Total} \\ \hline
Parameters & 4.2M & {4.2M} & {4.2M} & {4.2M} & {16.8 M}\\
\hline
\end{tabular}
\label{ta:np}
\end{table}
\vspace{-3mm}
\subsection{3D Pose based Action Recognition}
In order to compare the quality of the estimated 3D poses, two action recognition methods are used including a simple baseline CNN and an SGN~\cite{zhang2020semantics} based method.
In the baseline CNN method, as shown in Fig. \ref{fig:action}a, 
the predicted 3D poses are normalized using mean of all joints in a frame as $\mu(f)$ and magnitude of mean subtracted frame as $\sigma(f)$: 
\begin{equation}
\widehat{\textbf{p}}_3(f)=  \frac{1}{\sigma(f)}(\textbf{p}_3(f) -\mu(f)).
\end{equation}
Normalized poses are reshaped as 3D tensor where ($x$, $y$, $z$) coordinates are represented in the form of three channels which are scaled between $0-255$. 
This 3D tensor is again re-sized to  $224\times224\times3$ and passed through pre-trained ResNet50 to convert it into a feature vector of size $2048$ which is followed by a trainable fully connected layer of the same size as the number of action classes. A SoftMax layer is then used to convert the outputs to the class probabilities.

\begin{table}[b]
\small
\centering
\caption{Dataset Details: TV=Total Videos, TF=Total Frames,and MnF and MxF = Minimum  and Maximum Frames in any video.} 
\begin{tabular}{|c|c|c|c|c|c|c|c|}
\hline
Dset& TV & TF & MnF & MxF & Cams &Acts &Subj \\[0.5ex] 
\hline
H3.6M & 836 & 2.1M & 992 & 6343 &4& 15 & 7\\
SYSU & 480 & 0.1M & 58 & 638 & 1&12 & 40\\
NTU &  56880 & 4.9M & 32 & 300 &3& 60 & 40\\
\hline
\end{tabular}
\label{ta:de}
\end{table}
\begin{table*}[t]
\small
\centering
\caption{Mean Per Joint Position Error (MPJPE) comparison for 3D pose estimation with missing random joints on Human $3.6$M dataset. Baseline consists of Temporal Dilated Network without occlusion guidance.}
\setlength\tabcolsep{5pt} 
\begin{tabular}{c c c c c c c c c c c} 
\hline 
\hline
Method & None & Rand 2 & Rand 4 & Rand 6  & Rand 8 & Rand 10  & Rand 12 & Rand 14 & Rand 16 \\[0.5ex] 
\hline
\multicolumn{10}{c}{Protocol 1}\\
\hline
RN-Hier-Drop \cite{a02} & 59.7 & {65.9} & - & - & - & - & - & - & -  \\
Sem-GCN \cite{zhao2019semantic} & 42.1 & 490.6 & 714.6 & 833.4  & 887.5 & 912.7 & 904.1 & 883.5 & 856.5\\
VideoPose3D \cite{pavllo20193d} & \underline{37.5}  & {199.7} & {251.5} & {283.4} & {328.3} & {353.4} & {375.2} & {395.4} & {418.3}\\
Attention-3D \cite{liu2020attention} &  \textbf{34.7} & 743.6 & 987.4 & 1122.2 & 1198.2 & 1234.4 & 1245.0& 1097.5 & 1217.6 \\
\hline 
{Baseline} & {\underline{37.5}} &  {\underline{44.9}} &  {\underline{45.4}} & {\underline{48.0}} & {\underline{51.3}} & {\underline{58.9}} & {\underline{59.2}} & {\underline{66.3}} & {\underline{72.9}} \\
T3D CNN &  \underline{37.5} &  \textbf{36.3} &  \textbf{39.0} & \textbf{43.2} & \textbf{48.2} & \textbf{53.0} & \textbf{55.3} & \textbf{58.3} & \textbf{68.6}  \\
\hline 
\multicolumn{10}{c}{Protocol 2}\\
\hline
FConv \cite{moreno20173d} & 74.0 & 106.8 &- & -  & - & - & - & - & - \\
RN-Hier-Drop\cite{a02} & 45.6 & {51.0} & - & - & -&- & - & - & -\\
Sem-GCN \cite{zhao2019semantic} & 33.5 & 255.1 & 306.8 & 327.5 & 337.8 & 343.6 & 347.1 & 349.4 & 350.9\\
VideoPose3D \cite{pavllo20193d} & \underline{27.6}  & 118.8 & {154.8} & {177.2} & {207.2} & {221.8} & {232.6} & {238.6} & {235.8}\\
Attention-3D \cite{liu2020attention} &  \textbf{26.1} & 339.2 & 338.4 & 334.1 & 331.0& 328.8&327.2&325.3&325.4\\
\hline 

{Baseline} &  {\underline{27.6}} &  {\textbf{34.7}} &  {\textbf{35.0}} & {\textbf{37.0}} & {\textbf{39.2}} &{\textbf{44.4}} & {\textbf{44.7}} & {\textbf{49.8}} & {\textbf{55.4}}  \\
T3D CNN &  \underline{27.2} &  \textbf{28.4} &  \textbf{30.4} & \textbf{33.1} & \textbf{36.6} & \textbf{39.9} & \textbf{42.3} & \textbf{44.6} & \textbf{52.7} \\
\hline
\hline
\end{tabular}
\label{ta:2_2}
\end{table*}


{The second pose based action recognition method is motivated by Zhang \textit{et al.}~\cite{zhang2020semantics} as shown in Fig. \ref{fig:action}b.  It consists of an end-to-end network which utilizes joint-level information consisting of joint type as affinity matrix encoding human body structure, and frame-level information consisting of relations across multiple frames while maintaining the frame order. Further details of both of these methods are given in the supplementary material.}



\begin{table}
\small
\centering
\caption{MPJPE comparison for 3D pose estimation with missing fixed body parts on Human $3.6$M dataset.} 
\begin{tabular}{c c c c c c} 
\hline\hline
Methods &  L Arm & R Leg & Head & Lower Body  \\[0.5ex]
\hline
\multicolumn{5}{c}{Protocol 1} \\\hline
\hline
RN-Hier-Drop\cite{a02} &\underline{74.5} & \underline{70.4}&- & -  \\
Sem-GCN \cite{zhao2019semantic} & 391.2 & 278.0 & 430.2 & 216.4 \\
VideoPose3D \cite{pavllo20193d} & 310.5 &259.8 &\underline{221.7} &\underline{207.4}\\
Attention-3D \cite{liu2020attention} &  929.6 & 263.9 & 834.9 &979.7\\
\hline 
T3D CNN  &  \textbf{53.4} & \textbf{46.9} & \textbf{38.4} & \textbf{57.0}\\
\hline
\multicolumn{5}{c}{Protocol 2} \\\hline
FConv \cite{moreno20173d} & \underline{109.4} & \underline{100.2} & - &- \\
RN-Hier-Drop \cite{a02}& 63.0 & 55.2 & - & - \\
Sem-GCN \cite{zhao2019semantic} & 313.2 & 239.7 & 257.2 & 186.7\\
VideoPose3D \cite{pavllo20193d} & 217.2 & 193.7 & \underline{205.1} & \underline{163.0}\\
Attention-3D \cite{liu2020attention} &  358.8 & 197.9 & 289.5 & 262.2 \\
\hline 
T3D CNN &\textbf{49.0} & \textbf{36.4} & \textbf{29.9} & \textbf{47.2}\\[1ex]
\hline\hline 
\end{tabular}
\\
\label{ta:1_1}
\end{table}
\vspace{-2mm}
\begin{table}[b]
\small
\centering
\caption{MPJPE comparison for 3D pose estimation with complete occlusion scenarios: $t_1$, $t_3$, and $t_5$ shows one, three, and five consecutive occluded frames in Human 3.6M dataset.} 
\begin{tabular}{c c c c c c} 
\hline \hline
Methods & $t_1$ & $t_3$  & $t_5$\\[0.5ex]
\hline
\multicolumn{4}{c}{Protocol 1} \\\hline
VideoPose3D \cite{pavllo20193d} P1 & \underline{46.9} & \underline{49.3} & \underline{50.9}\\

Attention-3D \cite{liu2020attention} &  353.8 & 490.0 & 852.4 \\
T3D CNN & \textbf{38.5} &\textbf{38.4} & \textbf{39.6} \\
\hline
\multicolumn{4}{c}{Protocol 2} \\\hline
VideoPose3D \cite{pavllo20193d} & \underline{31.4} & \underline{32.2} & \underline{33.1}\\
Attention-3D \cite{liu2020attention}  & 266.1 & 362.8&285.8\\
T3D CNN & \textbf{28.8} &\textbf{28.8}& \textbf{29.3}\\

\hline \hline
\end{tabular}

\label{ta:3}
\end{table}
\section{Experiments and Results}
\label{sec:exp & res}
{A large number of experiments are performed to evaluate the proposed T3D CNN algorithm  on three publicly available datasets including Human $3.6$M~\cite{human3}, NTU RGB+D~\cite{shahroudy2016ntu} and SYSU 3D Human-Object Interaction dataset~\cite{hu2015jointly} (see Table \ref{ta:de}). We have compared our results with five existing state-of-the-art methods including static methods Rel-Hier-Drop \cite{a02}, Sem-GCN \cite{zhao2019semantic}, and FConv~\cite{moreno20173d} as well as the methods exploiting temporal information including VideoPose3D~\cite{pavllo20193d} and Attention-3D \cite{liu2020attention}. Sem-GCN~\cite{zhao2019semantic} uses Graph CNN and exploits semantic information to better estimate 3D poses. VideoPose3D~\cite{pavllo20193d} exploits temporal dilated convolution and Attention-3D \cite{liu2020attention} exploits attention mechanism to better handle occlusion cases. A baseline version of the proposed algorithm is also compared to evaluate the significance of the occlusion guidance mechanism. All results are reported for 243 frames temporal sequence length, unless stated otherwise.}

We use MPJPE for quantification of occlusion handling capability of different methods. It measures average error between joints in ground truth and in estimated 3D poses in millimeters (mm) after aligning root joints of both poses. 
We evaluate our framework using two protocols. In protocol 1, MPJPE is computed without any alignment while
in protocol 2, scaling, rotation and translation is applied to the predicted 3D pose to align it with the ground truth before calculating MPJPE. In addition to MPJPE we also employee action recognition accuracy as a measure of estimated 3D pose quality in the presence of occlusion. As the estimation quality degrades, the action recognition performance also degrades.
\subsection{Experiments on Human 3.6M Dataset}
It contains $3.6$ Million human poses for 15 different actions performed by 7 subjects and captured by 4 cameras. For evaluation, standard split is used: subjects \{1, 3, 5, 7\} are used for training and \{9, 11\} for testing. Occlusion handling capability of different methods is quantified by formulating partial and complete occlusion schemes. In partial occlusion, we formulate two different cases, random and fixed joints occlusion. In the first case, the number of randomly missing joints is varied as $\{2, 4, 8, 10, 12, 14, 16\}$, while in the second case left arm (3 joints), right leg (3 joints), head (1 joint) and lower body (4 joints) are missing. In complete occlusion case all joints are missing in $\{1, 3, 5\}$ consecutive frames.

Table \ref{ta:2_2} shows comparison of the proposed T3D CNN and Baseline with existing state-of-the-art methods \cite{pavllo20193d,zhao2019semantic,liu2020attention,a02} with random missing joints.  RN-hier-drop~\cite{a02} has provided results for random 2 missing, 
while for the remaining algorithms we have performed experiments using their pre-trained networks. As the number of missing joints increases, the error in all compared methods also increases, however T3D CNN has consistently remained the best performer for the estimation of 3D pose with missing 2D joints. 

Moreover, Gu \textit{et al.}~\cite{gu2021exploring} has reported 73.2mm MPJPE for 50\% random occlusion using 2D detections, which is significantly larger than 43.7mm obtained by T3D CNN.
Zhang \textit{et al.}~\cite{zhang2020object} have reported results with \{30\%, 70\%\} person occlusion which may be considered equivalent to random $\{6, 12\}$ missing joints. For both levels of occlusion, it obtained \{56.4, 68\} MPJPE which is again higher than  \{33.1, 56.4\} obtained by T3D CNN.
\begin{table*}
\small
\centering
\caption{MPJPE comparison for the 3D pose estimation using random missing joints on NTU RGB+D dataset. } 
\setlength\tabcolsep{5pt} 
\begin{tabular}{c c c c c c c c c c c c c c c c c c c} 
\hline 
\hline
Method & None & Rand 1 & Rand 4 & Rand 7  & Rand 13 & Rand 15  & Rand 18 & Rand 21 & Rand 23  \\[0.5ex] 
\hline
\multicolumn{10}{c}{Protocol 1}\\
\hline 
 VideoPose3D~\cite{pavllo20193d}  &  {115.6} &  {198.4} &  {315.7} &{372.9} & {435.0} & {446.5} & {459.6} & {467.9} & {471.6} \\
 Sem-GCN~\cite{zhao2019semantic} & 174.8 & 781.9 & 979.2 & 967.1 & 1080.1 & 1135.1 & 1218.4 & 1288.4 & 1329.8\\
 Attention-3D~\cite{liu2020attention} & \textbf{109.5} & 268.8 & 479.0 & 596.7 & 711.9 & 716.9 & 695.3 & 631.3 & 1704.3\\\hline
 {Baseline} & {\underline{115.6}} & {\underline{124.8}} & {\underline{140.7}} & {\underline{141.9}} & {\underline{148.2}} & {\underline{153.3}} & {\underline{159.2}} & {\underline{164.7}} & {\underline{164.9}} \\
 T3D CNN  &  \underline{115.6} & \textbf{124.5} & \textbf{140.4} & \textbf{140.9} & \textbf{146.2} & \textbf{148.2} & \textbf{147.8} & \textbf{152.5} & \textbf{154.6} \\
\hline 
\multicolumn{10}{c}{Protocol 2}\\
\hline
VideoPose3D~\cite{pavllo20193d} & {71.6} & {127.5} &  {184.6} & {184.6} & {239.2} & {246.3} & {255.5} & {263.5} & {268.3}\\
Sem-GCN~\cite{zhao2019semantic}& 100.1 & 187.1 & 251.2 & 281.7 & 313.3 & 319.7 & 327.3 & 333.1 & 336.2 \\
Attention-3D~\cite{liu2020attention} & \textbf{66.6} & 169.9 & 234.7 & 264.9 & 300.3 & 307.9 & 317.5 & 325.0 & 282.1 \\\hline
{Baseline} & {\underline{71.2}} & {\underline{76.1}} & {\underline{85.4}} & {\underline{86.8}}& {\underline{90.5}} & {\underline{92.9}} & {\underline{95.0}} &  {\underline{97.6}} & {\underline{96.9}}\\
T3D CNN &  \textbf{71.2} & \textbf{75.9} &  \textbf{85.1} &  \textbf{86.3} & \textbf{89.1} & \textbf{90.0} & \textbf{90.2} & \textbf{92.2} & \textbf{93.0} \\
\hline\hline
\end{tabular}
\label{ta:3_2}
\end{table*}
\begin{table}[b]
\small
\centering
\caption{Action recognition accuracy (\%) with  varying occlusion on NTU RGB+D dataset.} 
\begin{tabular}{|c|c c|c c|}
\hline
Pose Est.      & \multicolumn{2}{c|}{VideoPose3D} & \multicolumn{2}{c|}{T3D CNN} \\ \hline
Action Recog.  & CNN             & SGN            & CNN           & SGN          \\ \hline
Missing Joints & \multicolumn{4}{c|}{Accuracy \%}                                \\ \hline
0  &  61.33 & 76.27 & 61.33 & 76.27\\
1 & 42.91 &28.04 & 56.98 & 68.17  \\
4 & 25.02 & 6.31 & 52.98 & 62.33\\
7 &  17.86 & 4.02 & 51.51 & 60.00 \\
13 & 10.68 & 3.14 & 49.04 &55.98\\
15 & 9.07 & 2.99 & 48.37 & 55.12\\
18 & 8.32 & 2.77 & 48.19 &54.20\\
21 & 7.13 & 2.64 & 46.62 & 52.19\\
23 & 6.48 &2.56 & 45.86 & 50.96 \\
\hline \hline
\end{tabular}
\label{ta:5_1}
\end{table}

Table \ref{ta:1_1} shows the comparisons for fixed partial occlusion. Estimation improves by increasing the length of temporal sequence from 81 to 243 frames. It is because handling fixed missing joints is relatively easier than the random missing joints. Also an increase in the length of sequence increases the information about the action being performed. 
For the case of missing head estimation, error is minimum because given the position of shoulders, the head is within a small region. We have also experimented with complete occlusion case when 1, 3, and 5 consecutive frames are completely missing. We have compared our proposed method with \cite{pavllo20193d,liu2020attention} under Protocol 1 and 2 in Table \ref{ta:3}. 
In these experiments, the proposed framework was able to estimate good quality 3D poses despite severe occlusion, outperforming the existing methods by a significant margin.
\subsection{Experiments on NTU RGB+D Dataset}
NTU RGB+D is one of the largest skeleton dataset consisting of $56880$ videos performed by $40$ subjects across $60$ different action classes and three different views. As proposed by the existing state of the art methods, we used cross-subject evaluation in which $40$ subjects are equally divided into train and test splits. 
This dataset consists of 25 joint positions in 2D and 3D poses. 2D joints are input to the proposed T3D CNN while 3D joints are used as ground truth. We have evaluated proposed framework by randomly missing $\{1,4,7,13,15,18,21,23\}$ joints from each 2D pose.

Table \ref{ta:3_2} shows the comparison of the proposed approach with existing methods~\cite{pavllo20193d,zhao2019semantic,liu2020attention}. Attention-3D~\cite{liu2020attention} shows good performance when no joint is occluded but its performance degrades in case of missing joints. Sem-GCN \cite{zhao2019semantic} performance also drops due to occlusion because it is not utilizing temporal information. Our proposed T3D CNN shows consistently good performance even in heavily occluded scenarios such as missing 23 out of 25 joints. 

Table \ref{ta:5_1} shows the action classification accuracy of predicted 3D poses using CNN and SGN methods. Predicted 3D poses using the proposed CNN have shown improved classification accuracy under occlusion scenario whereas performance for predicted 3D poses by \cite{pavllo20193d} decreases much faster as the number of  missing joints increases. It shows the significance of our occlusion guided 3D pose estimation method.
\vspace{-4mm}
\subsection{Experiments on SYSU Dataset}
SYSU dataset consists of $480$ video sequences across 12 different actions performed by 40 different subjects. It contains skeletons with both 2D and 3D with 20 joint positions. 
In our setting half subjects are used for training and the others for testing. We have evaluated proposed framework with  $\{2, 4, 8, 12, 14, 16, 18\}$ random missing joints. 
Table \ref{ta:4_2} shows the MPJPE comparison of the proposed T3D CNN with existing state-of-the-art methods \cite{pavllo20193d,liu2020attention}. By increasing the number of random missing joints, error increases in all methods, however, T3D CNN has consistently shown much improved performance than the compared methods. { Experiments are also performed with OpenPose (OP)~\cite{op} 2D detections and confidence score as an occlusion guidance. We observe significant performance increase when occlusion guidance was used in T3D-OP compared to Basline-OP.}
Table \ref{ta:6_2} shows the action recognition accuracy using baseline CNN and SGN over the predicted 3D poses by VideoPose3D and proposed T3D CNN. In general, the action recognition score degrades as the number of missing joints increase, however, T3D CNN have consistently shown significantly improved scores due to effective occlusion handling. 

\begin{figure*}
\centering
\includegraphics[width=\textwidth]{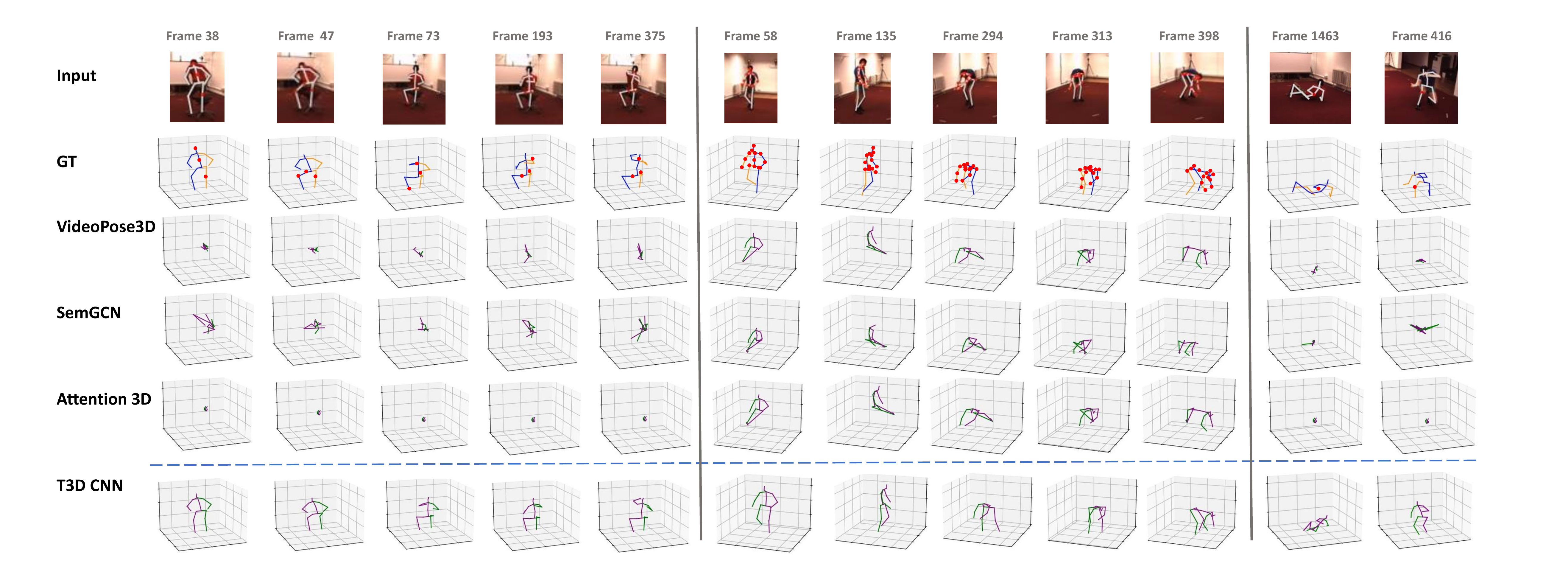}
\caption{Qualitative comparisons on Human $3.6$M dataset: (Left) 82\% random joints missing, (Middle)  lower body occluded and (Right) Failure cases: 97.2\% random joints miss with high error (non-occluded joints are shown by red markers).}
\label{fig1_all}
\end{figure*}

\begin{table*}
\small
\centering
\caption{MPJPE comparison for 3D pose estimation on SYSU dataset with random missing joints. {Baseline consists of Temporal Dilated Network without occlusion guidance module. Baseline-OP used Open-Pose~\cite{op} 2D detections while T3D-OP also used OpenPose confidence score for occlusion guidance.} }
\setlength\tabcolsep{5pt} 
\begin{tabular}{c c c c c c c c c c c c c c c c c c } 
\hline 
\hline
Method & None & Rand 2 & Rand 4 & Rand 8  & Rand 12 & Rand 14  & Rand 16 & Rand 18  \\[0.5ex] 
\hline
\multicolumn{9}{c}{Protocol 1}\\
\hline 
 VideoPose3D~\cite{pavllo20193d}  &  \underline{98.1} &  {181.2} &  {246.3} & {441.8} & {655.3} & {780.8} &{894.1} & {974.2}  \\
 Attention-3D~\cite{liu2020attention} & 111.8 & 361.9 & 590.7 & 574.9 & 883.6 & 1155.1 & 1427.9 & 1704.3\\\hline
 {Baseline-OP}  &  {112.6} & {116.9} & {125.7} & {128.8}  & {139.1} & {150.8} & {156.9}  & {166.1} \\
 
 {T3D-OP} & {112.6} & {114.4} & {116.3} & {119.3} & {120.8} & {125.3} & {132.8} & {142.6}\\\hline
 
 {Baseline}  &  {\textbf{94.8}} & {\underline{113.7}} & {\underline{121.7}} & {\underline{132.1}}  & {\underline{134.8}} & {\underline{145.8}} & {\underline{148.6}}  & {\underline{149.1}} \\

 T3D CNN  &  \textbf{94.8} & \textbf{110.0} & \textbf{114.9} & \textbf{120.9} & \textbf{123.1} & \textbf{125.9} & \textbf{125.4} & \textbf{131.8}  \\
\hline 
\multicolumn{9}{c}{Protocol 2}\\
\hline
VideoPose3D~\cite{pavllo20193d} & {52.0} & {102.4} & {141.9} & {201.1} & {240.8} & {254.4} & {266.2} & {272.4} \\
 Attention-3D~\cite{liu2020attention} & \textbf{49.9} & {121.8} & {183.3} & {233.4} & {255.7} &{264.7} & {273.7} & {282.1}\\\hline
  {BaselineOP}  & {66.9} & {70.0} & {71.2} & {74.2} & {76.5} & {83.6}  & {87.9} & {91.6}  \\
  {T3D conf} & {66.9} & {68.4} & {69.9} & {71.0} & {71.81} & {75.7} & {80.6} & {87.0}\\\hline
 {Baseline} &  {\underline{50.3}} & {\underline{58.7}} & {\underline{66.5}} & {\underline{67.7}} & {\underline{70.9}} & {\underline{73.3}}  & {\underline{76.3}} & {\underline{81.0}}  \\
{T3D CNN} & \underline{50.3} & \textbf{57.6} &  \textbf{60.0} &  \textbf{63.2} & \textbf{66.2} & \textbf{67.4} & \textbf{68.4} & \textbf{68.5} \\

\hline\hline
\end{tabular}
\label{ta:4_2}
\end{table*}

\begin{table}[t]
\small
\centering
\caption{Action recognition accuracy (\%) with  varying occlusion on SYSU dataset.} 
\begin{tabular}{|c|c c|c c|}
\hline
Pose Est.      & \multicolumn{2}{c|}{VideoPose3D} & \multicolumn{2}{c|}{T3D CNN} \\ \hline
Action Recog.  & CNN             & SGN            & CNN           & SGN          \\ \hline
Missing Joints & \multicolumn{4}{c|}{Accuracy \%}                                \\ \hline
0  &  62.32 & 98.24 & 62.32 & 98.24\\
2 & 20.96 &21.87 & 59.49 & 93.30 \\
4 & 19.83 & 17.86 & 59.49 & 91.51\\
8 &  12.75 & 9.82& 55.81 & 87.05\\
12 & 13.03 & 11.16 & 54.67 & 87.03\\
14 & 11.05 & 9.37 & 53.26 & 86.16\\
16 & 11.33 & 8.48& 51.56 & 82.14\\
18 & 11.61 & 8.92& 48.44 & 79.46\\ \hline
\end{tabular}\label{ta:6_2}
\end{table}

\subsection{Qualitative Comparisons}
Some visual results for missing joint estimation on Human $3.6$M  are presented in Fig. \ref{fig1_all} for fixed and random partial occlusion scenarios. Known joints are highlighted in red color. It can be seen that the proposed T3D CNN is able to estimate quite similar 3D poses as the ground truth, which demonstrates the effectiveness of T3D CNN while other compared methods including VideoPose3D, SemGCN, and Attention-3D have shown degraded performance for random partial occlusion and also these 
methods are unable to reconstruct lower body when it is occluded. Our proposed T3D CNN has performed quite well due to occlusion guidance that helps to identify missing  joints. Most of the visual results in Fig. \ref{fig1_all} demonstrate quite reasonable estimation of missing joints in addition to estimating the third dimension for all joints. We observe that our algorithm shows graceful degradation in case of complex actions. For example, in `Sitting Down' and `Photo' actions with 97.2 $\%$ occlusion, we obtain MPJPE of 117.8mm and 101.2mm respectively under protocol 1. Two frames with high error are shown in the last columns of Fig. \ref{fig1_all}. Some visual results on NTU RGB+D and SYSU  are shown in supplementary material.

\vspace{-4mm}
\begin{figure}[t]
\centering
\includegraphics[width=0.4\textwidth]{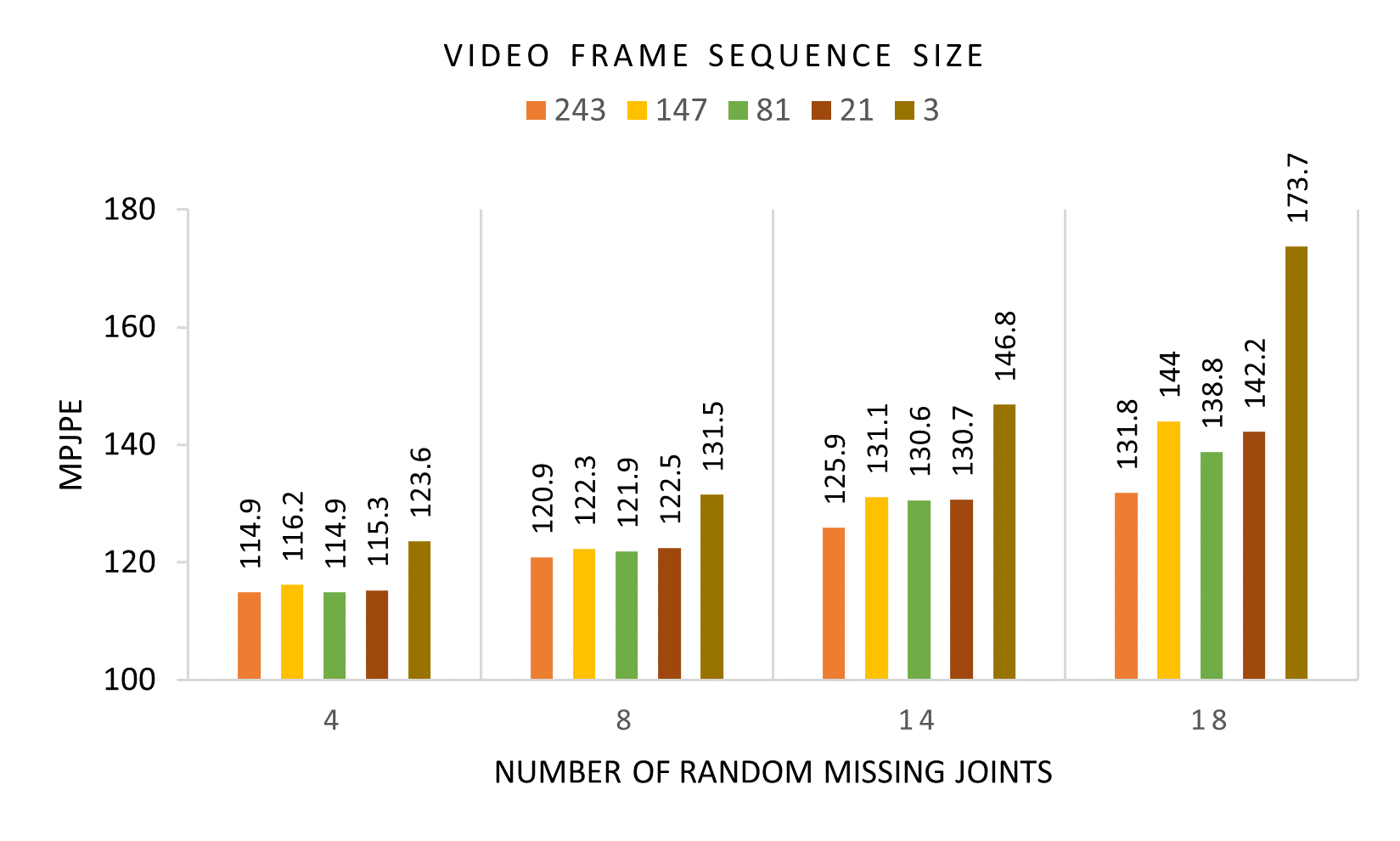}
\caption{3D pose estimation error (MPJPE) variation with varying temporal sequence size on SYSU dataset.}
\label{fig:bar}
\end{figure}
\subsection{Ablation Study}
\label{sec:abl}


To evaluate the significance of each component, different ablation studies are performed. Baseline results show the performance of proposed method using temporal sequence and without occlusion guidance mechanism. On Human $3.6$M (Table \ref{ta:2_2}) Baseline  observed  up to 8\% performance degradation compared to T3D CNN. Similarly, on NTU RGB+D Dataset (Table \ref{ta:3_2}) 12.2\% and on SYSU up to $23.2\%$ (Table \ref{ta:4_2}) degradation is observed. It shows the significant contribution of the proposed occlusion guidance mechanism in the performance.

{The occlusion guidance mechanism using confidence score obtained by OpenPose~\cite{op} is evaluated as `T3D-OP' as shown in Table \ref{ta:4_2} on SYSU dataset. For comparison, Baseline experiments using detected 2D pose has also been performed as `Baseline-OP'. T3D-OP obtained \{25.5\%, 24.1\%, 23.5\%\} improvement over Baseline-OP for \{14, 16, 18\} missing joints for Protocol 1. It also demonstrates the effectiveness of using confidence in the occlusion guidance module. Due to errors in the detected 2D poses by OpenPose compared to the ground truth values, direct comparison with T3D CNN may not be fair.}

{Performance variation is observed by varying the temporal sequence size as \{ 3, 21, 81, 147, 243 \} and experiment is repeated for \{ 4, 8, 14, 18 \} random missing joints on SYSU dataset (Fig. \ref{fig:bar}). Error difference between different sequence length increases as the number of missing joints increases. However, we consistently observe minimum error for the longest sequence of size 243 frames.}

\section{Conclusion}
\label{sec:con}
Estimation of 2D human pose from RGB images is quite mature and many pose detectors with excellent performance have been proposed. In the current work, a temporal dilated network based framework is proposed to estimate 3D human pose using 2D joint positions in the presence of missing joints. The input to the proposed framework is a temporal 2D pose sequence with occlusion guidance for missing joints and the output is  a  3D pose corresponding to the central pose in the input sequence. The proposed method efficiently maps data from 2D pose space to the 3D pose space utilizing the temporal information and occlusion guidance. In a large number of experiments, the proposed method outperformed  existing state-of-the-art methods using  random missing joints, fixed partial occlusion, and completely missing frames. As a future direction, one may integrate attention mechanism with our proposed occlusion guidance to better estimate 3D poses. Also, occlusion handling in the wild, occlusion in complex poses, and unsupervised occlusion handling are still open problems.  

\vspace{-5mm}
\bibliographystyle{IEEEtran}
\bibliography{t3d}
\vspace{-20mm}
\begin{IEEEbiography}[{\includegraphics[width=1in,height=1in,clip,keepaspectratio]{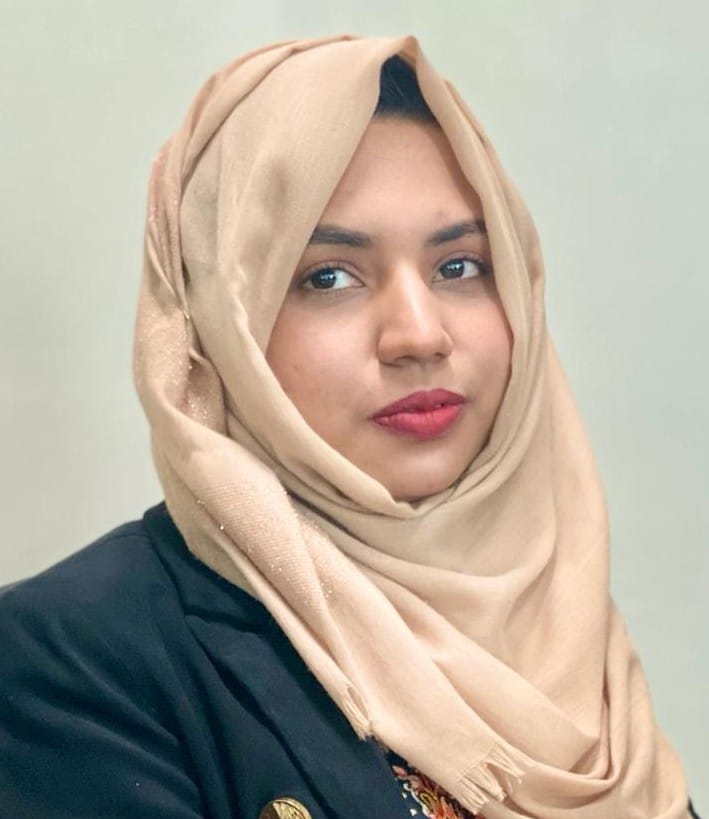}}]
{Mehwish Ghafoor}
is pursuing her PhD degree with the Computer Vision Lab at Information Technology University of the Punjab, Lahore. She has received her BS and MS degrees in Computer Science from Punjab University College of Information Technology, Lahore in 2014 and 2016, respectively. Her research interests include deep learning, human pose estimation, action recognition and segmentation.
\end{IEEEbiography}
\vspace{-20mm}
\begin{IEEEbiography}
	[{\includegraphics[width=1in,height=1in,clip,keepaspectratio]{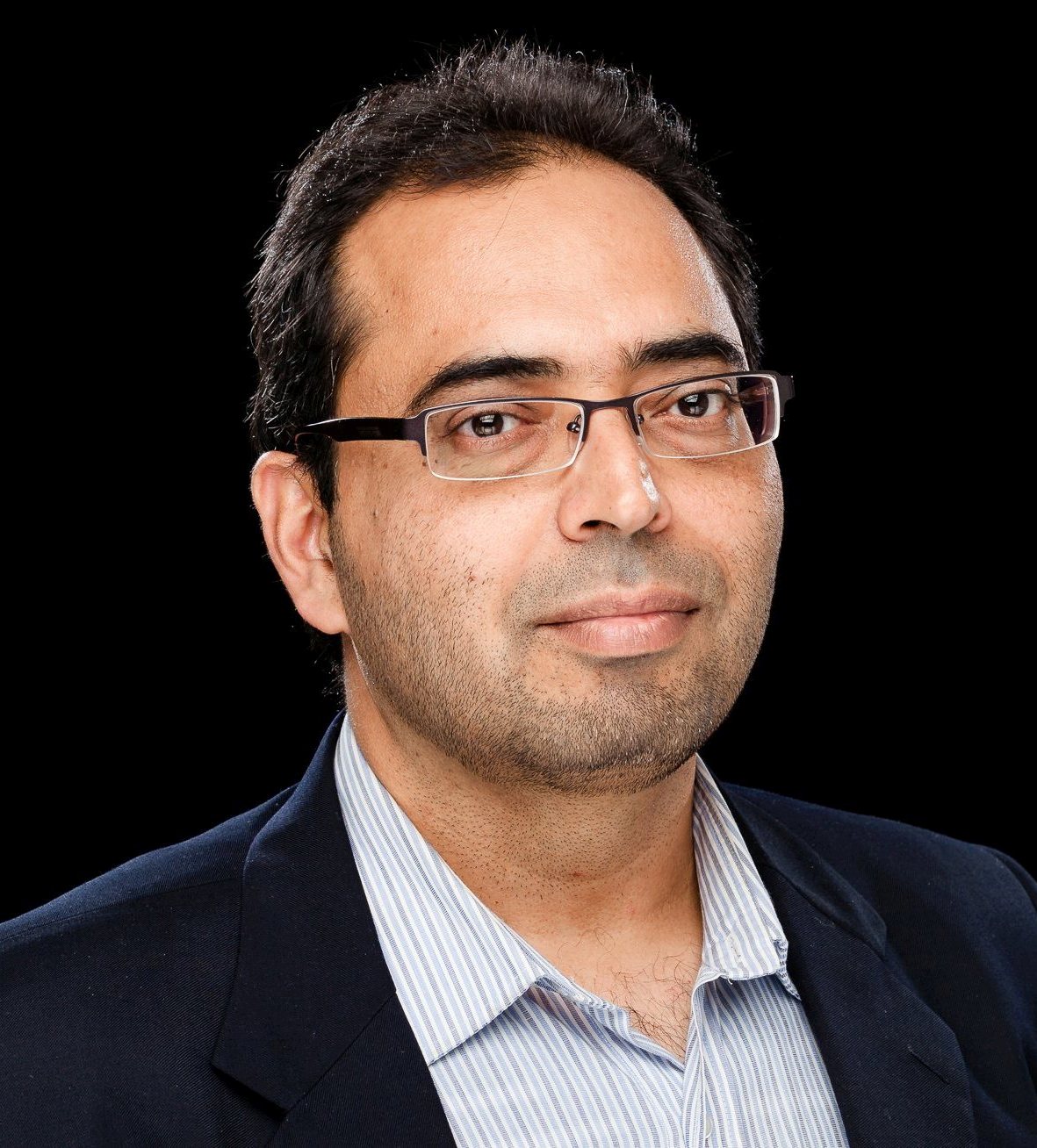}}]{Arif Mahmood} is a Professor and Chairperson of Computer Science Department in Information Technology University and Director Computer Vision Lab. His current research directions in Computer Vision are person pose detection and segmentation, crowd counting and flow detection, background-foreground modeling in complex scenes, object detection, human-object interaction detection and abnormal events detection. He is also actively working in diverse Machine Learning applications including  cancer grading and prognostication using histology images, predictive auto-scaling of services hosted on the cloud and the fog infrastructures, and environmental monitoring using remote sensing. He has also worked as a Research Assistant Professor with the School of Mathematics and Statistics, University of the Western Australia (UWA) where he worked on Complex Network Analysis. Before that he was a Research Assistant Professor with the School of Computer Science and Software Engineering, UWA and performed research on face recognition, object classification and action recognition. 
\end{IEEEbiography}
\vspace{-10mm}
\end{document}